\documentclass[11pt,a4paper]{article}
\usepackage{authblk}
\usepackage[hyperref]{emnlp-ijcnlp-2019}
\usepackage{times}
\usepackage{latexsym}

\usepackage{url}
\usepackage{graphicx}  
\usepackage{tabularx}
\usepackage{multirow}
\usepackage{booktabs}
\usepackage{tabu}
\usepackage[normalem]{ulem}
\usepackage[utf8]{inputenc}
\usepackage{algorithm}
\usepackage{algpseudocode}
\usepackage{array}
\usepackage{amsmath}
\usepackage{tablefootnote}

\usepackage{amsmath}
\usepackage{gb4e}

\makeatletter
\newcommand{\printfnsymbol}[1]{%
  \textsuperscript{\@fnsymbol{#1}}%
}
\makeatother

\noautomath 

\usepackage{lipsum}

\newcommand\blfootnote[1]{%
  \begingroup
  \renewcommand\thefootnote{}\footnote{#1}%
  \addtocounter{footnote}{-1}%
  \endgroup
}

\newcolumntype{L}[1]{>{\raggedright\let\newline\\\arraybackslash\hspace{0pt}}m{#1}}
\newcolumntype{C}[1]{>{\centering\let\newline\\\arraybackslash\hspace{0pt}}m{#1}}
\newcolumntype{R}[1]{>{\raggedleft\let\newline\\\arraybackslash\hspace{0pt}}m{#1}}

\aclfinalcopy 

\newcommand\confname{EMNLP-IJCNLP 2019}

\title{Instructions for \confname{} Proceedings}

\author[1]{\textbf{Paul Trichelair}\textsuperscript{*}}
\author[1]{\textbf{Ali Emami\textsuperscript{*}}}
\author[2]{\textbf{Adam Trischler}}
\author[2]{\textbf{Kaheer Suleman}}
\author[1]{\textbf{Jackie Chi Kit Cheung}}
\affil[1]{School of Computer Science, Mila/McGill University}
\affil[2]{Microsoft Research Montreal}
\affil[ ]{\textit {\{paul.trichelair, ali.emami\}@mail.mcgill.ca}}
\affil[ ]{\textit {\{adam.trischler, kasulema\}@microsoft.com}}
\affil[ ]{\textit {jcheung@cs.mcgill.ca}}

\date{}

\begin{document}
%
\title{How \textit{Reasonable} are Common-Sense Reasoning Tasks:\\ A Case-Study on the Winograd Schema Challenge and SWAG}

\maketitle

\begin{abstract}
 Recent studies have significantly improved the state-of-the-art on common-sense reasoning (CSR) benchmarks like the Winograd Schema Challenge (WSC) and SWAG. The question we ask in this paper is whether improved performance on these benchmarks represents genuine progress towards common-sense-enabled systems. We make case studies of both benchmarks and design protocols that clarify and qualify the results of previous work by analyzing threats to the validity of previous experimental designs. Our protocols account for several properties prevalent in common-sense benchmarks including size limitations, structural regularities, and variable instance difficulty. 
\end{abstract}
\blfootnote{*Equal contribution.}
\section{Introduction}
The proliferation of artificial-intelligence technologies that interact with human users (e.g., dialogue systems, recommendation systems, information retrieval tools) has led to renewed interest in common-sense reasoning (CSR). The progress of these technologies and the general societal reaction toward them greatly depend on advances in CSR, since systems can seem glaringly \emph{un}intelligent when they lack common sense.
Common sense is vital for resolving ambiguity that arises from implicit knowledge and under-specification. Consider the following sentence:
\begin{exe}
\ex \label{ex1} \normalsize The delivery truck zoomed by the school bus because \underline{it} was going so \textbf{fast}.

\end{exe}
Humans resolve the pronoun \textit{it} to \textit{the delivery truck} with no difficulty, whereas a system without common sense might be unable to distinguish the truck from the otherwise viable candidate, \textit{the school bus}. The above sentence is an example from a popular binary-choice pronoun co-reference problem called the Winograd Schema Challenge (WSC) \cite{levesque2011winograd}, designed to directly test a machine's grasp of common sense. What makes sentences like (\ref{ex1}) especially challenging for machine learning approaches is that they are formulated such that simple word co-occurrence statistics cannot resolve them at a rate above chance (i.e., \textit{the delivery truck} is unlikely to co-occur with \textit{going so fast} much more frequently than \textit{the school bus} does in large text corpora). In the same vein, a recently proposed common-sense inference task called SWAG \cite{zellers2018swag} further challenges co-occurrence-based approaches. SWAG's problem instances comprise a partial description, along with four candidate succeeding sentences designed to be distributionally similar. Among these, one successor is the most plausible. An example SWAG instance follows.

\begin{exe}
\ex \label{ex2} Someone is lifting the pinata. The pinata\\
a) drops from the swings.\\
b) bounces bigger than a third. \\
c) slumps across his shoulder back. \\
\textbf{d) falls on the ground.}

\end{exe}

\begin{table*}[h]
\centering
\begin{tabular*}{\linewidth}{p{0.12\textwidth}p{0.65\textwidth}p{0.05\textwidth}}
\toprule

Sentence Type & Examples & Proportion \\ \midrule

Non-Assoc.                & 
Bill passed the gameboy to John because \underline{his} turn was over. & 86.5\%
\\ \midrule
Assoc. & I'm sure that my map will show this \textbf{building}; \underline{it} is very \textbf{famous}. 
& 13.5\% \\

\bottomrule
\end{tabular*}
\caption{Examples and distribution of associative vs. non-associative WSC instances.}
\label{tab:examples}
\end{table*}

Recently, a number of systems have attained new state-of-the-art results on WSC and SWAG by querying a language model trained on a very large corpus \cite{trinh2018simple,radford2019lm,devlin2018bert}. The primary goal of this paper is to examine whether one can conclude from these systems' high performance on these CSR benchmarks that they actually possess common sense. We do so by systematically examining threats to the validity of experiments involving recent CSR models.

In particular, any study aiming to show a conclusion (e.g., that a particular model can perform CSR) is subject to threats to its internal and external validity \cite{campbell1963experimental}. Internal validity refers to whether the study is carried out correctly without any alternative explanations for its results, such as confounds or procedural errors. External validity refers to whether the results of the study can be generalized to other settings.

We find that most of the performance gains of recent approaches can be explained by issues with the experimental setup that concern validity threats of both types, but a small portion of those gains can be attributed to genuine progress. 

On WSC, the small size of the dataset and the predictable structure of its questions represent threats to external validity. We demonstrate this by applying perturbations to the dataset, whereby we switch the locations of the entities on a subset of data points. 
We find that the tested models' performances drop substantially in this new setting. 
We also analyze the portion of the performance gain not attributable to issues with the experimental setup in WSC, and find that current systems are very good at the subset of questions that require associative knowledge about semantic relatedness between words. 
Meanwhile, large parts of common-sense reasoning that require higher-level social, situational, or spatio-temporal awareness remain intractable.

In the case of SWAG, a possible confound is that the incorrect endings are generated semi-automatically by a language model, whereas the correct endings are generated by humans (threat to internal validity). We evaluate a representation model, BERT~\cite{devlin2018bert} on a modified version of the task that strips away the context sentence, such that models predict solely on the endings. We find that most (but not all) of the performance gain above chance level can be achieved by this deficient model.

\section{Related Work}
Our work presents new findings that reinforce realizations made in the community concerning the validity of a variety of different CSR tasks, most of which are in Natural Language Inference (NLI); some of these include that state-of-the-art models often do very well while being either agnostic to the premises in the task instance (which should be crucial for resolution) or by using linguistic cues that have little or nothing to do with world-knowledge or common-sense reasoning \cite{gururangan2018annotation}. 
In similar spirit, \citet{glockner2018} create an NLI test set specifically to show the deficiencies of state-of-the-art models in inferences that require lexical and world knowledge. Alternatively, validity checks through manual investigation as in \cite{Aikaterini2017} have revealed another NLI corpus to be vulnerable to errors and model exploitation. To the best of our knowledge, our work is the first analysis performed on two very popular CSR tasks, the WSC and SWAG, that have recently garnered considerable attention in the community and on which we are beginning to see models perform relatively well \cite{trinh2018simple,zellers2018swag}.

\section{Validity of CSR Experiments}
We now discuss the possible threats to the validity of CSR task setups in more detail.
\paragraph{Predictable Structure.} In general, instances from both WSC and SWAG exhibit distinctive regularities.
In SWAG, the counterfactual successor sentences are generated using an LSTM language model (LM), while the true successor comes from naturally occurring text. Despite recent advances in text generation, LSTM-generated responses feature stylistic patterns, such as repeated tokens, and display an overall lack of diversity \cite{xie2017neural}. The approach SWAG introduces to minimize stylistic artifacts, \textit{adversarial filtering}, is based on fooling a discriminator that classifies successors as human- or LM-generated. Nevertheless, upon inspecting the data, we found that LM-generated successors still contain repeated tokens and other signatures. A model that exploits these patterns could perform well without using any common sense.

An example regularity found in the WSC is that instances are often composed of two clauses connected by a causal discourse connective, like \textit{because} (as in (\ref{ex1})). This allows for simplifying assumptions~\cite{liu2016probabilistic} or schematizations~\cite{emami2018generalized}. The issue with exploiting these structural regularities is that systems become brittle to perturbations that would not affect the judgment of a human.

\paragraph{Limited Size.} Comprising only 273 test instances, the main drawback of the Winograd Schema Challenge is its limited size and the absence of training and validation sets for hyper-parameter tuning.
As a result, achieving above random accuracy on the WSC does not necessarily correspond to capturing common sense; it could be the result of a lucky draw.\footnote{The justification for this is included in the extra material.} 
\paragraph{Associativity.} The WSC task definition specifies that instances should not be resolvable via statistics that associate a candidate antecedent to other components of the sentence \cite{levesque2011winograd}. For example, in ``The lions ate the zebras because they are predators'' \cite{rahman2012resolving}, the pronoun \textit{they} can be resolved to \textit{lions} on the basis of a much stronger association of lions with predators than of zebras with predators. We will call this (flawed) type of instance \textit{associative} (or non-\textit{Google-proof} in \cite{levesque2011winograd}). Although the WSC should contain no associative sentences, there was no rigorous enforcement of this constraint. We therefore sought to quantify the associative proportion.
We only consider sentences to be associative if there is a clear argument for one antecedent being statistically preferred. Table \ref{tab:examples} outlines some examples and gives the associative proportions of the WSC.\footnote[2]{The details of the study can be found in the appendix. The related datasets are available at https://github.com/ptrichel/How-Reasonable-are-Common-Sense-Reasoning-Tasks}


\section{New Evaluation Protocols}
To probe the limitations discussed above, we propose evaluation protocols for the WSC and SWAG and apply them to several state-of-the-art methods. 

\paragraph{WSC.} First, we augment the existing dataset by switching candidates in sentences whenever possible (i.e., whenever switching the candidates does not obscure the sentence or affect the rationale to make the resolution decision). For example:
\begin{exe}
\ex \textbf{Original:} \textit{Emma} did not pass the ball to \textit{Janie} although \underline{she} saw that she was open.
\ex \textbf{Switched:} \textit{Janie} did not pass the ball to \textit{Emma} although \underline{she} saw that she was open.
\end{exe}
When switching the candidates \textit{Emma} and \textit{Janie}, the correct answer changes as well (from \textit{Emma} to \textit{Janie}). A system that relies on the entity itself to make a prediction produces the same answer when the candidates are switched, even though it should not. Thus, a system that correctly resolves both the original and the switched sentence can more confidently claim to reason about the full sentence, instead of exploiting a statistical quirk of the participant entities. We introduce two new metrics based on this observation: \textbf{accuracy on the switchable subset} before and after switching the candidates, and a \textbf{consistency score}.
The consistency score is the percentage of predictions that change (as would be expected) after candidates in the switchable subset are switched. 
In total, we counted 131 switchable instances in the WSC, which accounts for 47\% of the original problem set.\footnotemark[2]

\begin{table*}[h]
\begin{center}
\begin{tabu}to\linewidth{@{}X[l,3.5]X[c,2]X[c,2]X[c,2]X[c,2]@{}}
\toprule
Model                                   &  Full WSC Acc. & Unswitched Acc. & Switched Acc. & Consistency  \\ \midrule
Single LM            & 54.58\% & 54.96\% & 54.20\% & 56.49\% \\
Ensemble 10 LMs       & 61.54\% & 58.78\% & 49.62\% & 43.51\% \\
Ensemble 14 LMs       & 63.74\% & 63.36\% & 53.43\%  & 44.27\% \\
GPT-2 117M Full  & 55.68\% & 54.20\% & 54.20\% & 26.72\% \\
GPT-2 117M Partial  & 61.54\% & 59.54\% & 52.67\%  & 48.85\%\\
GPT-2 774M Full & 64.47\% & 62.60\% & 54.96\% & 45.04\% \\
GPT-2 774M Partial & 69.23\% & 67.94\% & 61.83\%  & 63.35\%\\
Knowledge Hunter    & 57.14\%\footnotemark[4] & 58.78\%\footnotemark[4] & 58.78\%\footnotemark[4]  & 90.07\%\footnotemark[4] \\
\bottomrule
\end{tabu}
\end{center}
\caption{Evaluation of state-of-the-art methods on WSC using the proposed switchability metrics. The last three columns give numbers on the switchable subset only.}
\label{table:evaluation}
\end{table*}
Taking special account of both the switchable and associative instances suggests the following evaluation protocol for a given model. First, we compute the accuracy on the original WSC and the accuracy on the \textit{switchable} subset of the WSC before and after switching the candidates, and compute the corresponding consistency score. Next, we compute the accuracy on the \textit{associative} subset. A model can be tailored to use statistical information about the entities but perform poorly when this cannot be exploited.
\paragraph{SWAG.} When evaluating on SWAG, it is important to determine whether the prediction relies on an understanding of the context or on shallow patterns in the LM-generated counterfactuals. To isolate this effect, we remove the context from the problem instances, keeping only the four successors. Three of these are machine generated. Predicting the correct label thus amounts to discriminating the human-written successor from the machine-generated ones. By comparing the performance difference between a model that has access to the context versus one that does not, we can determine the extent to which the model actually relies on contextual reasoning. 

\section{Experiments}
We test several recently proposed systems using our proposed protocols: specially-trained, ensembled language models (LMs) \cite{trinh2018simple}, a large language model GPT-2 \cite{radford2019lm} a knowledge hunting method \cite{emami2018generalized}; and a fine-tuned representation model, BERT \cite{devlin2018bert} for SWAG.\footnote[3]{We include implementation details in the appendix.}

In both \citet{trinh2018simple} and \citet{radford2019lm}, the language model scores the two sentences obtained when replacing the pronoun by the two candidates.
The sentence that is assigned a higher probability designates the chosen candidate. Probability is calculated via the chain rule, as the product of the probabilities of each word in the sentence. The knowledge hunting method, from \citet{emami2018generalized}, is a rule-based system that uses search engines to gather evidence for the candidate resolutions without relying on the entities themselves. BERT, a pre-trained deep bidirectional representation, is fine-tuned for SWAG using a softmax over the four possible endings.
\begin{table}
\begin{center}
\begin{tabu}to\linewidth{@{}X[4,l]X[2.25,r]X[2.25,r]@{}}

\toprule
Model       & Assoc.    &  Non-Assoc.    \\ \midrule
Single LM                & 73.0\%     & 51.7\% \\
Ensemble 10 LMs               & 91.9\% &  56.8\% \\
Ensemble 14 LMs     &   83.8\% & 60.6\% \\ 
GPT-2 117M Full & 73.0\% & 53.0\% \\
GPT-2 117M Partial & 78.4\% & 58.9\%\\
GPT-2 774M Full & 81.1\% & 61.9\% \\
GPT-2 774M Partial & 91.9\%  & 65.7\%\\
Knowledge Hunter     & 50.0\%\footnotemark[4]&  58.3\%\footnotemark[4] \\
\bottomrule
\end{tabu}
\caption{Accuracy of state-of-the-art methods on associative and non-associative WSC instances.}
\label{table:googleproof}
\end{center}
\end{table}

\section{Results} 
\paragraph{WSC.} Performance of the  state-of-the-art methods with respect to our proposed switchability metrics is shown in Table \ref{table:evaluation}. We observe that accuracy is stable across the different subsets for the single LM and GPT-2 117M with full scoring.  However, the performance of the ensembled LMs and GPT-2 117M with partial scoring  falls back to near random on the switched subset. This correlates with a lower consistency score and suggests that the two models overfit to the dataset. The GPT-2 774M language models, the largest available ones, show the highest accuracy on the WSC, despite a significant drop in accuracy on the switched subset. In addition, they show the highest consistency scores on the WSC. As for the knowledge hunting method,
it performed relatively well on the entire WSC, and is 100\% consistent by definition, since it does not utilize the entities themselves during resolution. 
\footnotetext[4]{This is the expected accuracy and consistency. For those instances that the knowledge hunter did not acquire evidence, we expect half to be correct by chance.}
\begin{table}

\begin{center}
\begin{tabu}to\linewidth{@{}X[2.5,l]X[2,r]@{}}
\toprule
Discriminator Model       & Accuracy   \\ \midrule
Successor-only              & 70.0\% \\
Full model & 80.9\% \\
\bottomrule
\end{tabu}
\caption{Evaluation of BERT on SWAG using the proposed metrics.}
\label{table:evaluationSWAG}
\end{center}
\end{table}

In Table~\ref{table:googleproof}, we present model performance on the associative and non-associative subsets of the WSC. 
These demonstrate that LM-based methods perform very well on the associative sentences, as expected. 
However, their performance drops significantly on the non-associative subset, when information related to the candidates themselves does not give away the answer.
\paragraph{SWAG.} The performance of the state-of-the-art model is shown in Table \ref{table:evaluationSWAG}. We observe that the model can distinguish human and LM-generated endings with an accuracy of 70.0\%. This suggests that a strong performance on SWAG can be obtained without any consideration of the context, and that the task may not be well-suited to evaluate CSR. Nevertheless, BERT performs at 10.9\% above this score when it uses the full context, indicating that the model does possess some ``understanding'' of the described situation.

\section{Conclusion}
The function of common sense in AI systems is both important and difficult to address. This paper is an attempt to make experiments, namely those performed on the WSC and SWAG, more rigorous by examining threats to the validity of these experimental designs. Based on the protocols we introduce, we show that performing at state-of-the-art on these datasets does not necessarily imply strong common-sense reasoning capability. 
We are happy to see a rising interest in the WSC in the community, including very recent work by \citet{ruan2019exploring} and \citet{sap2019socialiqa}, which reinforces the need for proper evaluation protocols. With the release of an increasing number of fine-grained inference tasks aimed at these abilities \cite{roemmele2011choice,morgenstern2016planning,wang2018glue,rashkin2018event2mind,mccann2018natural}, the issue of experimental validity in CSR will also become even more important. 

\section*{Acknowledgements}
This work is supported by the Natural Sciences and Engineering Research Council of Canada, and by the Canada CIFAR AI Chair program. The first authors are supported in part by Microsoft Research.

\bibliography{emnlp-ijcnlp-2019}

\begin{thebibliography}{20}
\expandafter\ifx\csname natexlab\endcsname\relax\def\natexlab#1{#1}\fi

\bibitem[{Campbell and Stanley(1963)}]{campbell1963experimental}
Donald~T Campbell and Julian~C Stanley. 1963.
\newblock \emph{Experimental and quasi-experimental designs for research}.
\newblock Houghton Mifflin.

\bibitem[{Devlin et~al.(2018)Devlin, Chang, Lee, and
  Toutanova}]{devlin2018bert}
Jacob Devlin, Ming-Wei Chang, Kenton Lee, and Kristina Toutanova. 2018.
\newblock Bert: Pre-training of deep bidirectional transformers for language
  understanding.
\newblock In \emph{NAACL-HLT}.

\bibitem[{Emami et~al.(2018)Emami, De~La~Cruz, Trischler, Suleman, and
  Cheung}]{emami2018generalized}
Ali Emami, Noelia De~La~Cruz, Adam Trischler, Kaheer Suleman, and Jackie
  Chi~Kit Cheung. 2018.
\newblock A knowledge hunting framework for common sense reasoning.
\newblock In \emph{Proceedings of the Conference on Empirical Methods in
  Natural Language Processing}.

\bibitem[{Glockner et~al.(2018)Glockner, Shwartz, and Goldberg}]{glockner2018}
Max Glockner, Vered Shwartz, and Yoav Goldberg. 2018.
\newblock Breaking nli systems with sentences that require simple lexical
  inferences.
\newblock In \emph{Proceedings of the 56th Annual Meeting of the Association
  for Computational Linguistics (Volume 2: Short Papers)}, page 650–655.
  Association for Computational Linguistics.

\bibitem[{Gururangan et~al.(2018)Gururangan, Swayamdipta, Levy, Schwartz,
  Bowman, and A.~Smith}]{gururangan2018annotation}
Suchin Gururangan, Swabha Swayamdipta, Omer Levy, Roy Schwartz, Samuel Bowman,
  and Noah A.~Smith. 2018.
\newblock Annotation artifacts in natural language inference data.
\newblock In \emph{Proceedings of the 2018 Conference of the North American
  Chapter of the Association for Computational Linguistics: Human Language
  Technologies, Volume 2 (Short Papers)}, pages 107--112. Association for
  Computational Linguistics.

\bibitem[{Kalouli et~al.(2017)Kalouli, Real, and de~Paiva}]{Aikaterini2017}
Aikaterini-Lida Kalouli, Livy Real, and Valeria de~Paiva. 2017.
\newblock Textual inference: getting logic from humans.
\newblock In \emph{Proceedings of the 12th International Conference on
  Computational Semantics (IWCS)}.

\bibitem[{Levesque et~al.(2011)Levesque, Davis, and
  Morgenstern}]{levesque2011winograd}
Hector~J Levesque, Ernest Davis, and Leora Morgenstern. 2011.
\newblock The winograd schema challenge.
\newblock In \emph{AAAI Spring Symposium: Logical Formalizations of Commonsense
  Reasoning}, volume~46, page~47.

\bibitem[{Liu et~al.(2016)Liu, Jiang, Evdokimov, Ling, Zhu, Wei, and
  Hu}]{liu2016probabilistic}
Quan Liu, Hui Jiang, Andrew Evdokimov, Zhen-Hua Ling, Xiaodan Zhu, Si~Wei, and
  Yu~Hu. 2016.
\newblock Probabilistic reasoning via deep learning: Neural association models.
\newblock \emph{arXiv preprint arXiv:1603.07704}.

\bibitem[{McCann et~al.(2018)McCann, Keskar, Xiong, and
  Socher}]{mccann2018natural}
Bryan McCann, Nitish~Shirish Keskar, Caiming Xiong, and Richard Socher. 2018.
\newblock The natural language decathlon: Multitask learning as question
  answering.
\newblock \emph{arXiv preprint arXiv:1806.08730}.

\bibitem[{Morgenstern et~al.(2016)Morgenstern, Davis, and
  Ortiz~Jr}]{morgenstern2016planning}
Leora Morgenstern, Ernest Davis, and Charles~L Ortiz~Jr. 2016.
\newblock Planning, executing, and evaluating the winograd schema challenge.
\newblock \emph{AI Magazine}.

\bibitem[{Radford et~al.(2019)Radford, Wu, Child, Luan, Amodei, and
  Sutskever}]{radford2019lm}
Alec Radford, Jeffrey Wu, Rewon Child, David Luan, Dario Amodei, and Ilya
  Sutskever. 2019.
\newblock Language models are unsupervised multitask learners.
\newblock \emph{cloudfront preprint:d4mucfpksywv}.

\bibitem[{Rahman and Ng(2012)}]{rahman2012resolving}
Altaf Rahman and Vincent Ng. 2012.
\newblock Resolving complex cases of definite pronouns: the winograd schema
  challenge.
\newblock In \emph{Proceedings of the 2012 Joint Conference on Empirical
  Methods in Natural Language Processing and Computational Natural Language
  Learning}, pages 777--789. Association for Computational Linguistics.

\bibitem[{Rashkin et~al.(2018)Rashkin, Sap, Allaway, Smith, and
  Choi}]{rashkin2018event2mind}
Hannah Rashkin, Maarten Sap, Emily Allaway, Noah~A Smith, and Yejin Choi. 2018.
\newblock Event2mind: Commonsense inference on events, intents, and reactions.
\newblock In \emph{Proceedings of the 56th Annual Meeting of the Association
  for Computational Linguistics}.

\bibitem[{Roemmele et~al.(2011)Roemmele, Bejan, and
  Gordon}]{roemmele2011choice}
Melissa Roemmele, Cosmin~Adrian Bejan, and Andrew~S Gordon. 2011.
\newblock Choice of plausible alternatives: An evaluation of commonsense causal
  reasoning.
\newblock In \emph{AAAI Spring Symposium: Logical Formalizations of Commonsense
  Reasoning}.

\bibitem[{Ruan et~al.(2019)Ruan, Zhu, Ling, Shi, Liu, and
  Wei}]{ruan2019exploring}
Yu-Ping Ruan, Xiaodan Zhu, Zhen-Hua Ling, Zhan Shi, Quan Liu, and Si~Wei. 2019.
\newblock Exploring unsupervised pretraining and sentence structure modelling
  for winograd schema challenge.
\newblock \emph{arXiv preprint arXiv:1904.09705}.

\bibitem[{Sap et~al.(2019)Sap, Rashkin, Chen, LeBras, and
  Choi}]{sap2019socialiqa}
Maarten Sap, Hannah Rashkin, Derek Chen, Ronan LeBras, and Yejin Choi. 2019.
\newblock Socialiqa: Commonsense reasoning about social interactions.
\newblock \emph{arXiv preprint arXiv:1904.09728}.

\bibitem[{Trinh and Le(2018)}]{trinh2018simple}
Trieu~H Trinh and Quoc~V Le. 2018.
\newblock A simple method for commonsense reasoning.
\newblock \emph{arXiv preprint arXiv:1806.02847}.

\bibitem[{Wang et~al.(2018)Wang, Singh, Michael, Hill, Levy, and
  Bowman}]{wang2018glue}
Alex Wang, Amanpreet Singh, Julian Michael, Felix Hill, Omer Levy, and Samuel
  Bowman. 2018.
\newblock Glue: A multi-task benchmark and analysis platform for natural
  language understanding.
\newblock In \emph{Proceedings of the Conference on Empirical Methods in
  Natural Language Processing}.

\bibitem[{Xie(2017)}]{xie2017neural}
Ziang Xie. 2017.
\newblock Neural text generation: A practical guide.
\newblock \emph{arXiv preprint arXiv:1711.09534}.

\bibitem[{Zellers et~al.(2018)Zellers, Bisk, Schwartz, and
  Choi}]{zellers2018swag}
Rowan Zellers, Yonatan Bisk, Roy Schwartz, and Yejin Choi. 2018.
\newblock Swag: A large-scale adversarial dataset for grounded commonsense
  inference.
\newblock \emph{arXiv preprint arXiv:1808.05326}.

\end{thebibliography}
\bibliographystyle{acl_natbib}
\newpage
\appendix 
\section{Dataset construction}
\subsection{Switching candidates}
This dataset contains the original WSC with the switched version of each sentence whenever the process does not obscure the sentence or affect the rationale used to resolve the target pronoun. To construct this dataset, we first automatically switch the two candidates.
\begin{exe}
\ex \textbf{Original sentence} \textit{Emma} did not pass the ball to \textit{Janie} although \underline{she} saw that she was open.
\ex \textbf{Switched sentence} \textit{Janie} did not pass the ball to \textit{Emma} although \underline{she} saw that she was open.
\end{exe}

This process can make a sentence obscure, as in the following example:
\begin{exe}
\ex \textbf{Original sentence} Sam broke both his \textit{ankles} and he's walking with \textit{crutches}. But a month or so from now \underline{they} should be better.
\ex \textbf{Switched sentence} Sam broke both his \textit{crutches} and he's walking with \textit{ankles}. But a month or so from now \underline{they} should be better.
\end{exe}

The sentence obtained is not correct as \textit{walking with ankles} is neither semantically correct nor requires the same resolution rationale. To filter out these sentences, we asked three English native speakers, who did not have prior knowledge on the WSC, to classify the sentences as \textit{Switchable} or \textit{Not Switchable}. We keep the switched version of the sentence if the three annotators agreed. This procedure produces a dataset of 131 switched sentences with a high agreement as shown in Table \ref{table:agreement}.
\begin{table*}[ht]
\begin{center}
\begin{tabu}to\linewidth{@{}X[1.5,l]X[2.75,r]X[2.75,r]@{}}
\toprule
Statistic used  & Score Switchability     & Score Associativity  \\ \midrule
Fleiss' Kappa     & 0.96 & 0.79\\
\end{tabu}
\caption{Inter-rater agreement measured using Fleiss's Kappa for both the switching and the associativity annotations}
\label{table:agreement}
\end{center}
\end{table*}
\subsection{\textit{Associativity}}

This dataset contains the original WSC sentences labeled as \textit{associative} or \textit{non-associative}. Associative Winograd sentences are those in which one candidate antecedent associates strongly with the clause containing the pronoun, while the other candidate antecedent exhibits no such association strength. For example:

\begin{exe}
\ex In the storm, \textit{the tree} fell down and crashed through \textit{the roof} of my house. Now, I have to get [it] repaired.
\end{exe}

Here, \textit{the roof} can be argued to be much more strongly associated with \textit{repaired}, and on this basis, can be used to resolve the pronoun.

An example of a non-associative sentence is: 

\begin{exe}
\ex Everyone really loved \textit{the oatmeal cookies}; only a few people liked \textit{the chocolate chip cookies}. Next time, we should make more of [them] .
\end{exe}

Here, we don't expect, at least \textit{a priori}, that \textit{oatmeal cookies} associate more than \textit{the chocolate chip cookies} with the clause, \textit{"we should make more of them"} and therefore can be argued to be much more robust to techniques that rely on co-occurence statistics. 

We split the WSC into smaller associative and non-associatve datasets by conducting a human study similar to that in A.1. The three annotators only had access to the clause containing the pronoun (e.g. \textit{get [it] repaired} and \textit{Next time, we should make more of [them]} for (5) and (6) respectively), and the two candidate antecedents. Using these, they were asked to categorize a sentence as associative or non-associative according to whether or not they saw a strong association between one entity and the clause, and no such association with the other entity. We chose to consider a sentence as \textit{associative} if the three annotators unanimously agreed. This process lead to a high inter-annotator agreement as shown in Table \ref{table:agreement} and resulted in an \textit{associative} dataset with 37 sentences and a \textit{non-associative} dataset with 252 sentences (there were 42 sentences for which there was not a full agreement).

\section{Lucky draw}
We consider a random classifier so that for each sentence, it chooses one of the two candidates. Since the dataset is balanced, the probability of getting the correct answer is 50\%. When classifying the 273 instances, the number of correct answers $X$ is a binomial random variable. The probability of getting more than 55\% accuracy (more than 150 correct answers) is given by:

\begin{align*}
    P(X>150)&=1-P(X\leq150)\\
    P(X>150)&=1-\sum_{i=0}^{150}P(X=i)\\
    P(X>150)&=1-\sum_{i=0}^{150}{{273}\choose{i}}0.5^i(1-0.5)^{273-i}\\
    P(X>150)&=1-0.5^{273}\sum_{i=0}^{150}{{273}\choose{i}}\\
    P(X>150)&=0.04\\
\end{align*}
It shows that the probability of scoring more than 55\% on the WSC using a random classifier is 4\%. When repeating the experiments 10 times, the probability that one of the experiments gives an accuracy greater than 55\% corresponds to $1-P(X\leq150)^{10}=0.37$. Practically, on the WSC, this means that if we have a pool of 10 random classifiers, there is more than a 1-in-3 chance that one of them scores more than 55\%.

\section{Implementation Details}
For the WSC, we reproduced the results for the language model and the Knowledge-Hunter using the authors' released code available on Github\footnote{The language model: \url{https://github.com/tensorflow/models/tree/master\\/research/lm\_commonsense}}
\footnote{The Knowledge Hunter: \url{https://github.com/aemami1/Wino-Knowledge-Hunter}}.\\
For GPT-2, we use the implementation released in the paper and slighly modified it. We have attached the implementation with the submission.\\
For BERT, we have attached the implementation with the submission. The modifications we have made to the original implementation include the necessary adapations for SWAG.

\end{document}